\documentclass[conference,a4paper] {IEEEtran}
\IEEEoverridecommandlockouts
\usepackage{cite}
\usepackage{amsmath,amssymb,amsfonts}
\usepackage{algorithmic}
\usepackage{graphicx}
\usepackage{booktabs}
\usepackage{adjustbox}
\usepackage{multirow} 
\usepackage{textcomp}
\usepackage{xcolor}
\usepackage{times}
\usepackage{amsmath}
\usepackage{amssymb}
\usepackage{multicol,multirow}
\usepackage[tableposition=top]{caption}
\usepackage{cite}
\usepackage{booktabs}
\usepackage{bigstrut}
\usepackage{floatrow}
\usepackage{rotating}
\usepackage{adjustbox}
\usepackage{amsmath}
\usepackage{url}
\def\BibTeX{{\rm B\kern-.05em{\sc i\kern-.025em b}\kern-.08em
    T\kern-.1667em\lower.7ex\hbox{E}\kern-.125emX}}
\begin{document}

%%%%%%%%% TITLE
\title{LFRA-Net: A Lightweight Focal and Region-Aware Attention Network for Retinal Vessel Segmentation}

\author{%
\IEEEauthorblockN{Mehwish Mehmood$^{1,*}$, Shahzaib Iqbal$^{2}$, Tariq Mahmood Khan$^{3}$, Ivor Spence$^{1}$, Muhammad Fahim$^{1}$}
\IEEEauthorblockA{$^{1}$ \textit{School of Electronics, Electrical Engineering and Computer Science, Queen's University Belfast, United Kingdom}\\
$^{2}$ \textit{Department of Electrical Engineering}, Abasyn University Islamabad Campus (AUIC), Islamabad, Pakistan\\
$^{3}$ \textit{Department of Cybersecurity and Digital Forensics}, Naif Arab University for Security Sciences, Riyadh, Saudi Arabia}
\textit{Emails: mmehmood01@qub.ac.uk, shahzeb.iqbal@abasynisb.edu.pk, tkhan@nauss.edu.sa, \{i.spence, m.fahim\}@qub.ac.uk}}

\maketitle
\begingroup
\renewcommand{\thefootnote}{\fnsymbol{footnote}}
\footnotetext[1]{Corresponding author}
\endgroup
\thispagestyle{empty}
%%%%%%%%% ABSTRACT
\begin{abstract}
Retinal vessel segmentation is critical for the early diagnosis of vision-threatening and systemic diseases, especially in real-world clinical settings with limited computational resources. Although significant improvements have been made in deep learning-based segmentation methods, current models still face challenges in extracting tiny vessels and suffer from high computational costs. In this study, we present LFRA-Net by incorporating focal modulation attention at the encoder-decoder bottleneck and region-aware attention in the selective skip connections. LFRA-Net is a lightweight network optimized for precise and effective retinal vascular segmentation. It enhances feature representation and regional focus by efficiently capturing local and global dependencies. LFRA-Net outperformed many state-of-the-art models while maintaining lightweight characteristics with only 0.17 million parameters, 0.66 MB memory size, and 10.50 GFLOPs. We validated it on three publicly available datasets: DRIVE, STARE, and CHASE\_DB. It performed better in terms of Dice score (84.28\%, 88.44\%, and 85.50\%) and Jaccard index (72.86\%, 79.31\%, and 74.70\%) on the DRIVE, STARE, and CHASE\_DB datasets, respectively. LFRA-Net provides an ideal ratio between segmentation accuracy and computational cost compared to existing deep learning methods, which makes it suitable for real-time clinical applications in areas with limited resources. The code can be found at https://github.com/Mehwish4593/LFRA-Net.
\end{abstract}

%%%%%%%%% BODY TEXT

%% Keywords

% \end{frontmatter}

%% \linenumbers % Uncomment if line numbers are required

\section{Introduction}
\IEEEPARstart{R}{etinal} image analysis using computer technologies for disease diagnosis is an emerging area of research. Retinal images are often used to diagnose blood vessel-related diseases as well as other diseases like diabetic retinopathy (DR), neurodegenerative disorders, glaucoma, age-related macular degeneration (AMD), multiple sclerosis (MS) and cardiovascular disease \cite{soomro2016automatic, khan2019boosting, soomro2018impact, khan2019generalized, khan2019ggm}. Diagnosis of these diseases without computer-aided technologies requires specialized knowledge,  time, commitment, and financial resources, and the probability of errors is significant\cite{iqbal2023robust, khan2023feature, abbasi2023lmbis, iqbal2023ldmres, khan2024lmbf, khan2024esdmr, xu2025edge}. To ensure early prevention or treatment, robust retinal image segmentation is essential for early disease screening \cite{khawaja2019improved, khawaja2019multi, khan2019use, khan2020residual, khan2020shallow}. Deep learning can potentially transform autonomous disease detection in the medical field \cite{khan2020exploiting, khan2020semantically, naveed2021towards, khan2022width, khan2021rc, khan2022t, iqbal2022g}. Recent research emphasizes the importance of retinal vascular segmentation in detecting symptoms of neurodegenerative diseases, including dementia. 60–80\% of dementia cases are caused by Alzheimer's disease \cite{khan2025role}. Significant changes in retinal vessels have been observed in the patients with Alzheimer's disease \cite{javeed2023machine, arsalan2022prompt, brahmavar2023ikd+, khan2023retinal}. \\

In this research, we present LFRA-Net, a new lightweight segmentation model designed especially for effective segmentation of retinal blood vessels. We have trained and evaluated our model on three publicly available datasets: DRIVE, STARE, and CHASE\_DB. Metrics such as the Dice score, the Jaccard index, the sensitivity, and the specificity are used to evaluate the performance of our model. Additionally, a trainable parameter count, FLOPs and memory size are used to evaluate the model's complexity. The effectiveness of our model is validated by comparing performance metrics with other state-of-the-art segmentation models. The primary contributions of this paper are as follows.\\

\begin{enumerate}
    \item We introduce LFRA-Net, a lightweight retinal vascular segmentation network that offers both accuracy and computational complexity, making it ideal for resource-constrained clinical contexts.\\
    
    \item We implement focal modulation attention at the encoder-decoder bottleneck to enhance crucial feature representation and convey comprehensive information for efficient segmentation. This increases contextual awareness of vessel structures.\\
    
   \item We use region-aware attention in selective skip connections to improve spatial context representation without increasing model complexity. We also present a comprehensive ablation that provides architectural insights and demonstrates that LFRA-Net outperforms larger and more complex models, making it ideal for clinical deployment in the real world.

\end{enumerate}

\section{Literature Review}
Retinal image segmentation has shown remarkable efficiency using DL-based U-shaped encoder-decoder architectures. Using skip connections and symmetric network structures to capture local and global details efficiently, traditional approaches such as U-Net \cite{ronneberger2015u}, DeconvNet \cite{noh2015learning}, RefineNet \cite{lin2017refinenet} and Mask R-CNN \cite{he2017mask} established the way for subsequent innovations. Numerous changes and lightweight variations have been proposed to enhance the basic U-Net. For instance, Jingfei et~al. proposed Salient U-Net \cite{8842560}, a bridge-like saliency mechanism in a U-Net architecture that incorporates foreground objects and uses a cascade technique. S-UNet successfully resolves the data imbalance and improves retinal vascular segmentation using a bridge-like topology and a cascade of prominent features. Miu et~al. \cite{Liu2023} presented Wave-Net, a pixel-wise retinal vessel extraction approach for medical image segmentation. To get high accuracy, this model uses multi-scale feature fusion (MFF) and a detailed enhancement and denoising (DED) block in place of the fundamental skip connection from U-Net. Similarly, J. Li et~al. \cite{Li2023} presented the multi-scale attention-guided fusion network, which uses multi-scale attention blocks, attention-guided fusion, hybrid feature pooling, and feature improvement for retinal vascular segmentation. Despite its remarkable accuracy and F1 score, it has a high inference time.\\

M3U-CDVAE \cite{Yu2023} is a thin refinement network explicitly designed for retinal vascular segmentation presented by Yang Yu et~al. It works in three stages: pre-segmentation, segmentation, and refining. It uses the first 13 layers of MobileNet-V3 as the encoder backbone. This model is suitable for single-task learning because it balances high accuracy and F1 scores with fewer parameters. Liu et~al. \cite{Liu20231} presented residual depth-wise over-parameterized (ResDO)-UNet, which combines ResDO-conv with multiple pooling operations and pooling-attention fusion blocks to enable multi-scale feature fusion and non-linear pooling for improved segmentation. Wentao et~al. \cite{liu2022full} presented FR-UNet, a segmentation method intended to increase segmentation accuracy and vessel connectivity. This technique uses a multi-resolution convolution mechanism to preserve full image resolution and a dual-threshold iterative approach to capture weak vessel pixels. However, it tends to produce false positives, especially when thin vessel segmentation is involved. In order to segment and categorize retinal vessels, Chowdhury et~al. \cite{chowdhury2022msganet} created MSGANet-RAV, a U-shaped encoder-decoder network. Although it had difficulty correctly segmenting smaller vessels and complex structures, this model performed well in handling vessel crossings. Lyu et~al. \cite{Lyu2022} introduced a convolutional neural network (CNN) model for fractal dimension evaluation and binary vessel segmentation, which produced results similar to U-Net but had problems when it came to managing pathological variation in medical images. To provide comprehensive information, Xu et~al. \cite{XU2022695} suggested a dual-channel asymmetric CNN that integrates segmentation results from both channels based on pre-processing scale and orientation features. However, despite its advantages, the technique uses more GPU memory because it is affected by differences in vascular shape and disease characteristics.\\

Many researchers have created lightweight networks especially for medical image segmentation. In medical imaging, achieving excellent performance with low model complexity and fast inference is still tricky. Tarasiewicz et~al. \cite{tarasiewicz2020lightweight} trained several tiny networks across all image channels using multimodal magnetic resonance imaging (MRI) to construct lightweight U-Nets that accurately identify brain cancers. By substituting pyramidal convolution for all convolutional layers in traditional U-Net, PyConvU-Net \cite{li2021pyconvu} improves the segmentation accuracy using fewer parameters. However, PyConvU-Net's inference time is still insufficient. TBConvL-Net \cite{iqbal2025tbconvl}, PLVS-Net \cite{arsalan2022prompt}, Lmbf-net \cite{khan2024lmbf}, G-Net Light \cite{iqbal2022g}, and MKIS-Net \cite{khan2022mkis} are CNN architectures that are lightweight and efficient for segmenting retinal blood vessels. The current lightweight approaches have two significant shortcomings. First, their performance is lower than that of the state-of-the-art methods, and second, they are unable to effectively generalize to unseen data. Bhati et~al. \cite{Bhati2025} introduced DyStA-RetNet, a shallow encoder-decoder CNN with attention blocks. This model combines multi-scale dynamic attention with a statistical spatial attention module in the encoder and a partial decoder to retrieve high- and low-level information. The described gating refinement increases segmentation performance by highlighting vessels of varying widths while suppressing unnecessary details. Abbasi et~al. \cite{Abbasi2024} introduced the LMBiS-Net, a “Lightweight Multipath Bidirectional Skip” CNN designed for retinal vessel segmentation. It incorporates multipath feature extraction blocks to capture vessel information at different levels and bidirectional skip connections two ways from the encoder to the decoder for better cross-flow of information. With DCNet, a dilated convolution network comprising only three convolutional layers, Shang et~al. \cite{Shang2024} further simplified the approach. Each of the layers contains dilated kernels, which provide multi-scale contextual capture, realizing large receptive fields spatially without deep stacks.

\section{Methodology}
We have developed a lightweight model with an encoder-decoder architecture for the segmentation of retinal blood vessels. The following section discusses the architecture of our proposed model. 
\begin{figure*}[h]
  \centering
    
    \includegraphics[width=\textwidth]{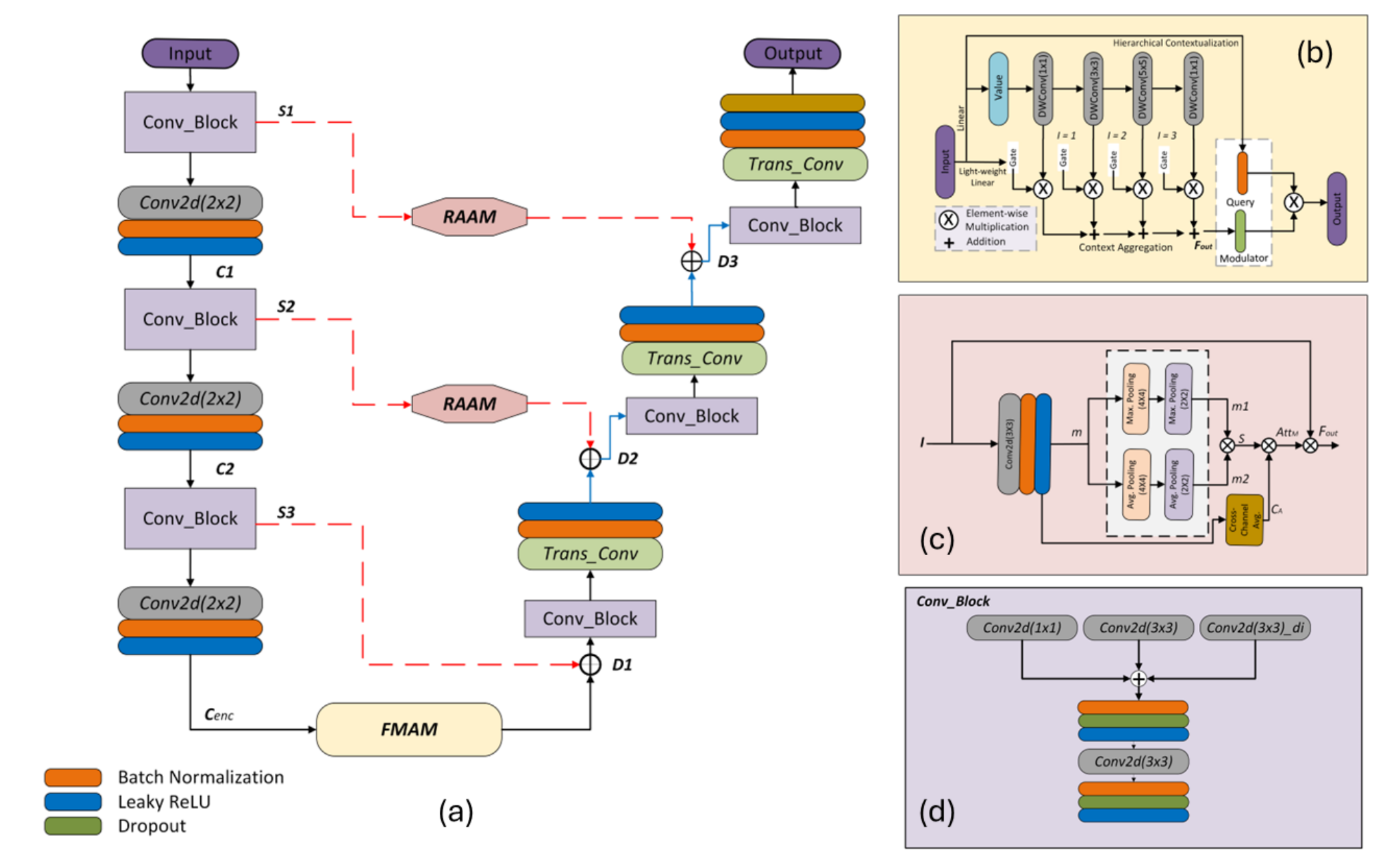}  
    \caption{Architecture of the proposed LFRA-Net: (a) Focal Modulation Attention Mechanism with context aggregation (FMAM); (b) Region Aware Attention Mechanism (RAAM); (c) Convolutional Block.}\label{LFRA-Net}
\end{figure*}

\subsection{Model Architecture}\label{subsec:architecture}
The proposed model design, LFRA-Net, is presented in Figure \ref{LFRA-Net}(a). It introduces a lightweight encoder-decoder architecture designed to overcome the limitations of segmentation models in capturing small vessel structures or relying on computationally expensive architectures. It efficiently captures spatial and contextual features for vessel segmentation. The architecture consists of three primary components: multiscale convolution blocks, Region-Aware Attention Machinism (RAAM)\cite{naveed2024ra}, and Focal Modulation Attention Machinism (FMAM)\cite{yang2022focal} for feature enhancement. Skip connections integrate encoder outputs into the decoder and preserve multiscale information, ensuring that crucial input features are preserved. Multiscale convolution blocks are implemented in the encoder for the extraction of multiple features from the input. The input image \textit{I} with a resolution of 512 $\times$ 512 pixels, is first passed through a stack of multiscale convolution blocks implemented in the encoder as shown in Figure \ref{LFRA-Net}(d). These blocks combine standard and dilated convolutions to extract features across receptive fields. As a result, LFRA-Net can effectively capture both local vessel borders and more general contextual information without adding more parameters or deepening the model.
\begin{equation}
  C_{ms} =    C^{1\times 1}(\mathit{I})\oplus C^{3\times 3}(\mathit{I})\oplus C_{di}^{3\times 3}(\mathit{I})
  \label{eq1}
\end{equation}
\begin{equation}
  C_1 =   \mathit{LeakyReLU}(Dr^{(0.5)}({BN}(C_{ms})))
  \label{eq2}
\end{equation}
\begin{equation}
  MS\_Conv(\mathit{I)}= \mathit{LeakyReLU}(Dr^{(0.5)}({BN}(C^{3\times3}(C_1))))
  \label{eq3}
\end{equation}
Each convolution block combines $C^{1\times 1}$, $C^{3\times 3}$ and $C_{di}^{3\times 3}$, which denote the convolution with kernel sizes of 1$\times$1, 3$\times$3, and 3$\times$3 dilated convolutions, respectively, to capture diverse spatial and contextual features. These multiscale features are combined and processed through a series of operations, including dropout with a probability of 0.5 ($Dr^{(0.5)}$), leaky ReLU activation (\textit{LeakyReLU}), and batch normalization (\textit{BN}). The output is further processed using a convolutional layer with a 3$\times$3 kernel size ($C^{3\times 3}$), Leaky ReLU, batch normalization, and dropout for further refinement. Multiscale convolution blocks are also employed in the decoder during the upsampling stages to produce high-resolution outputs gradually. Furthermore, RAAM is incorporated into first and second skip connections, which connect the encoder to the upsampling blocks of the decoder to improve contextual information and fine-tune spatial features. It improves early high-resolution features while optimizing computation. Applying attention to later skip connections offers fewer benefits since they are less spatially detailed. FMAM is included in the encoder-decoder architecture's bottleneck to improve feature representations.\\

The encoder comprises three multiscale convolution blocks, each followed by a down-sampling operation using a \(2 \times 2\) convolution. These blocks include a series of convolutional layers with batch normalization, leaky ReLU activation, and dropout for feature extraction. The first skip connection (S1) is obtained in Eq. \ref{eq4}.
\begin{equation}
  S1 = MS\_Conv(\mathit{I})
  \label{eq4}
\end{equation}
The initial encoder block employs the Leaky ReLU activation function and batch normalization after a $2\times2$ convolution operation ($C^{2\times 2}$) on \textit{S1} for downsampling. The output of the initial convolutional block is given in Eq. \ref{eq5}.
\begin{equation}
 \mathit{C1} = \mathit{LeakyReLU}({BN}(C^{2\times2}(S1)))
  \label{eq5}
\end{equation}
The second skip connection employs a multiscale convolution block to the resulting initial convolutional block ($C1$) as shown in Eq. \ref{eq6}.
\begin{equation}
 S2 = MS\_Conv(\mathit{C1})
\label{eq6}
\end{equation}
The second encoder block employs the Leaky ReLU activation function and batch normalization after a $2 \times 2$ convolution operation ($C^{2 \times 2}$) on $S2$. The output of the second encoder block is given in Eq. \ref{eq7}.
\begin{equation}
 \mathit{C2} = \mathit{LeakyReLU}({BN}(C^{2\times2}(S2)))
  \label{eq7}
\end{equation}
Similarly, the outputs of the third skip connection and encoder block are given in Eqs. \ref{eq8} and \ref{eq9}, respectively.
\begin{equation}
 S3 = MS\_Conv(\mathit{C2})
\label{eq8}
\end{equation}
\begin{equation}
 \mathit{C_{enc}} = \mathit{LeakyReLU}({BN}(C^{2\times2}(S3)))
  \label{eq9}
\end{equation}

Skip connections from the encoder are concatenated with the up-sampled features to preserve fine-grained details. Each decoder block applies a transposed convolution followed by a multiscale convolution block for feature refinement. The initial decoder block is obtained by applying FMAM to the final encoder block ($C_{enc}$,) as shown in Eq. \ref{eq10}.

\begin{equation}
  \mathit{D1} = \mathcal{F}(C_{enc})
  \label{eq10}
\end{equation}

where FMAM is denoted as $\mathcal{F}$. The second decoder block is obtained by applying a multiscale convolution block to \textit{D1} followed by transpose convolution, batch normalization, and a leaky ReLU activation function. The resulting feature map is then concatenated with the second skip connection, which incorporates RAAM to enhance feature fusion, as shown in Eq. \ref{eq11}.

\begin{equation}
   \mathit{D2} = \mathcal{R}(S2) \oplus \mathit{LeakyReLU}(\mathit{BN} (T^{3\times 3}(MS\_Conv(D1))))
   \label{eq11}
\end{equation}

where \textit{T} and $ \mathcal{R}$ represent the transpose convolution and RAAM, respectively. Similarly, the output of the third decoder block is given in Eq. \ref{eq12}.  
\begin{equation}
   \mathit{D3} =  \mathcal{R}(S1) \oplus\mathit{LeakyReLU}(\mathit{BN} (T^{3\times 3}(MS\_Conv(D2))))
   \label{eq12}
\end{equation}
 The final decoder block generates the predicted output through a \(1 \times 1\) convolution, followed by a sigmoid activation function, given in Eq. \ref{eq13}:

\begin{equation}
   I_{out} = \sigma(C^{1 \times 1}(MS\_Conv(D3)))
   \label{eq13}
\end{equation}
where \(\sigma\) represents the sigmoid function.\\

The Region-Aware and Focal Modulation Attention are two attention modules combined to improve feature representation even more. In order to enhance delicate vessel structures and maintain spatial detail, particularly in early high-resolution layers, RAAM is applied to the skip connections. In order to enhance global feature modulation and ensure better context before decoding, FMAM is implemented at the bottleneck. We chose these modules with attention to improve segmentation accuracy while maintaining a computationally light architecture. The following sections provide details of these attention modules.
\subsection{Focal Modulation Attention Mechanism}\label{subsec:FMAM}
To enhance the extracted feature information, FMAM is integrated between the encoder and the decoder. As illustrated in Fig. \ref{LFRA-Net}(b), FMAM comprises three primary components. It begins by processing the encoder’s output to encode visual details across both short and long ranges using a sequence of depth-wise convolutional layers. Each layer in the stack extracts hierarchical feature representations, capturing both local and global information. The output of each layer is mathematically represented as:
\begin{equation}
z^{\mathit{(l)}} = \text{DWConv}(z^{\mathit{(l-1)}}),
\end{equation}
where \(z^{\mathit{(l)}}\) denotes the output feature map at level \(\mathit{l}\), and \(\text{DWConv}\) represents a depth-wise convolution applied at the same level.

To capture the global context, an average pooling operation is applied to the final level of the depth-wise convolutions, given by:

\begin{equation}
z^{(L+1)} = \text{AvgPool}(z^{(L)}),
\end{equation}

where \(z^{(L)}\) represents the feature map at the final depth-wise convolution level and \(z^{(L+1)}\) is the globally aggregated feature.

The mechanism further refines these features through hierarchical context aggregation and modulation. The aggregated features across all levels are combined as:

\begin{equation}
Z_{\text{out}} = \sum_{\mathit{l}=1}^{L+1} G^{\mathit{(l)}} \odot z^{\mathit{(l)}},
\end{equation}
where \(G^{\mathit{(l)}}\) represents the gating weights at level \(\mathit{l}\), and \(\odot\) denotes element-wise multiplication.
The final modulated output for a query token \(F_i\) is computed as:
\begin{equation}
F_i = q(C_{enc(i)}) \odot h(Z_{\text{out}}),
\end{equation}
where \(i\) is the spatial index of the feature map, \(q(C_{enc(i)})\) is the query projection function, and \(h(Z_{\text{out}})\) is the modulator projection function. 
\subsection{Region-Aware Attention Mechanism (RAAM)}\label{subsec:raam}
The RAAM enhances spatial and contextual feature learning by applying an attention mechanism to the encoder output. It is illustrated in Figure \ref{LFRA-Net}(c).
Suppose that a model's input tensor is $I$, which is passed through the operations mentioned in Eq. \ref{eq14} to get the output m.
\begin{equation}
 m = ReLU(BN(C^{3 \times 3}(I)))
  \label{eq14}
\end{equation}
where the convolution layer is represented by \(C\), and the batch normalization is symbolized as \(BN\). The average and max pooling are then employed to \(m\) to compute the features (Eqs. \ref{eq15} and \ref{eq16}. These pooling operations of the specified tensor are combined using a multiplication operation as shown in Eq. \ref{eq17}.
\begin{equation}
m1 = M_p^{2 \times2}M_p^{4\times4}(m)),
 \label{eq15}
\end{equation}
\begin{equation}
m2 = A_p^{2 \times2}A_p^{4\times4}(m)),
 \label{eq16}
\end{equation}
\begin{equation}
S = m1 \otimes m2
 \label{eq17}
\end{equation}
where \(\otimes\) represents element-wise multiplication. \(M_p^{2\times2}\) and \(A_p^{2\times2}\) denote max pooling and average pooling with filter sizes of \(2 \times 2\) and \(M_p^{4\times4}\) and \(A_p^{4\times4}\) denote max pooling and average pooling with filter sizes of \(4 \times 4\) , respectively. 
The semantic feature map is computed using cross-channel averaging maps as follows:
\begin{equation}
C_A = \frac{1}{M} \sum_{j=1}^{M} m_{i,j} \quad (i \in \{1, 2, \ldots, N\})
\label{eq18}
\end{equation}

Attention maps (\(Att_M\)) are computed as:
\begin{equation}
Att_M = \frac{1}{N} \sum_{i=1}^{N} S_i C_{A(i)}
\label{eq19}
\end{equation}
The number of feature channels required for each pixel to have differentiating areas is represented by \(M = 16\) the number of classes (N = 2). Finally, Eq. \ref{eq20} modifies the input tensor I by using attention maps:
\begin{equation}
F_{out} = I \otimes Att_M
\label{eq20}
\end{equation}
where \(F_{out}\) is the final output feature tensor.\\

\subsection{Loss Function and Optimization}\label{subsec:loss}
The proposed model is trained using a weighted dice loss function to handle class imbalances effectively. The dice loss is defined as:

\begin{equation}
   \mathcal{L}_{dice} = 1 - \frac{2 |S \cap G|}{|S| + |G|},
\end{equation}

where \(S\) and \(G\) denote the segmented output and ground truth, respectively. The model is optimized using the Adam optimizer, ensuring robust convergence. Adam was selected because of its capacity to adaptively modify learning rates while training, which is particularly useful for segmenting thin and tiny structures like retinal arteries.

\section{Experiments and Results}\label{Results}
\subsection{Datasets and implementation}\label{subsec:datasets}
We tested our model using the DRIVE, STARE, and CHASE\_DB datasets. The DRIVE dataset \cite{qureshi2013manually} contains forty retinal images, each with $565\times 584$ pixels of resolution. The STARE dataset \cite{STAREDataset} has 20 color fundus images with a resolution of $700 \times 605$. The CHASE\_DB dataset \cite{7530915} is composed of 28 images, each representing $1024\times 1024$ pixels of resolution. Table \ref{tab:datasets} shows the detailed insight into the datasets. Each dataset was evaluated using the standard train/test split (20/20 for DRIVE, 16/4 for STARE, and 20/8 for CHASE\_DB). Reported results are based on designated test sets to ensure fair comparison and prevent overfitting. We used 80\% and 20\% of the images for model training and validation from each training dataset with a batch size of 8 and an ADAM optimizer with a learning rate of 0.002.

\begin{table}[!h]
\centering
\caption{Overview of datasets and their properties, including the number of training and testing images, total and augmented images, original image resolution, field of view (FOV), and training details, providing better insight into their application.}
\label{tab:datasets}
\adjustbox{max width=\textwidth}{
\begin{tabular}{lccc}
\toprule
\textbf{Property} & \textbf{DRIVE}~\cite{qureshi2013manually} & \textbf{STARE}~\cite{STAREDataset} & \textbf{CHASE\_DB}~\cite{7530915} \\
\midrule
Training Images & 20 & 16 & 20 \\
Testing Images & 20 & 4 & 8 \\
Total Images & 40 & 20 & 28 \\
Augmented Images & 1080 & 1024 & 1080 \\
Resolution (pixels) & 565$\times$584 & 700$\times$605 & 1024$\times$1024 \\
Resized to & 512 & 512 & 512 \\
Field of View (FOV) & 35 & 45 & 45 \\
\bottomrule
\end{tabular}
}
\end{table}
\begin{table}[!h]
\centering
\caption{Computational complexity comparison between LFRA-Net and other state-of-the-art methods.}
\label{complex_results}
\adjustbox{max width=\textwidth}{
\begin{tabular}{lccc}
\toprule
\textbf{Method} & \textbf{Params (M)} & \textbf{FLOPs (G)} & \textbf{Size (MB)} \\
\midrule
U-Net~\cite{ronneberger2015u}         & 7.76  & 96.68   & 29.60 \\
U-Net++~\cite{zhou2018unet++}         & 9.04  & 238.52  & 34.49 \\
Attn U-Net~\cite{oktay2018attention}  & 9.25  & 371.68  & 35.33 \\
IterNet~\cite{li2020iternet}         & 13.60 & 194.40  & 94.70 \\
GT-DLA-dsHFF~\cite{yuan2021multi}    & 26.00 & 473.90  & 2.30  \\
LiViT-Net~\cite{tong2024livit}       & 6.90  & 71.10   & 27.60 \\
FS-u-net~\cite{jiang2024retinal}      & 0.87  & 47.60   & 3.50  \\
\textbf{LFRA-Net}                     & \textbf{0.17} & \textbf{10.50} & \textbf{0.66} \\
\bottomrule
\end{tabular}}
\end{table}
\begin{table*}[!h]
  \centering
  \caption{Performance comparison of LFRA-Net with recent methods on the DRIVE, STARE, and CHASE\_DB datasets. The performance measures are highlighted, with the best results emphasized in bold for clarity.}
  \label{tab:Vessels}
  \adjustbox{max width=\textwidth}{
\begin{tabular}{lccccccccccccc}
    \toprule
    \multirow{3}{*}{\textbf{Method}} & \multirow{3}{*}{\textbf{Param (M)}} & \multicolumn{12}{c}{\textbf{Performance Measures in (\%)}} \\
     & & \multicolumn{4}{c}{\textbf{DRIVE}} & \multicolumn{4}{c}{\textbf{STARE}} & \multicolumn{4}{c}{\textbf{CHASE\_DB}} \\
    \cmidrule(lr){3-6}\cmidrule(lr){7-10}\cmidrule(lr){11-14}
     & & \textbf{Dice} & \textbf{J} & \textbf{Sn} & \textbf{Sp} & \textbf{Dice} & \textbf{J} & \textbf{Sn} & \textbf{Sp} & \textbf{Dice} & \textbf{J} & \textbf{Sn} & \textbf{Sp} \\
    \midrule
    BCD-UNet \cite{azad2019bi}      & 20.65 & 82.49 & 69.33 & 79.84 & 98.03 & 82.30 & 68.14 & 78.92 & 98.16 & 79.32 & 67.42 & 77.35 & 98.01 \\
    U-Net++ \cite{zhou2018unet++}   & 9.04  & 80.60 & 68.27 & 78.40 & 98.00 & 81.40 & 69.02 & 79.02 & 98.36 & 83.49 & 66.88 & 82.83 & 98.21 \\
    Att. U-Net \cite{oktay2018attention} & 9.25  & 80.39 & 67.21 & 79.06 & 98.31 & 81.06 & 68.39 & 78.04 & \textbf{98.87} & 79.64 & 66.17 & \textbf{84.84} & 98.31 \\
    U-Net \cite{ronneberger2015u}   & 7.76  & 81.41 & 68.64 & 80.57 & 98.33 & 81.18 & 68.56 & 70.50 & 98.84 & 78.98 & 65.26 & 76.50 & \textbf{98.84} \\
    MultiRes-UNet \cite{ibtehaz2020multiresunet} & 7.20  & 82.32 & 69.26 & 79.46 & 97.89 & 82.44 & 68.27 & 77.09 & 98.48 & 80.12 & 67.09 & 80.10 & 98.04 \\
    FR\_UNet \cite{liu2022full}     & 5.72  & 83.16 & 71.20 & \textbf{83.56} & 98.37 & 83.30 & 72.46 & 83.27 & 98.69 & 81.51 & 68.82 & 87.98 & 98.14 \\
    SegNet \cite{badrinarayanan2017segnet} & 1.42  & 83.02 & 70.23 & 80.18 & 98.26 & 83.41 & 70.71 & 80.12 & 98.65 & 81.96 & 68.56 & 81.38 & 98.24 \\
    IterNet \cite{li2020iternet}     & 13.60 & 82.18 & 69.21 & 77.35 & 98.38 & 81.46 & 68.76 & 77.15 & 98.86 & 80.73 & 67.44 & 79.70 & 98.20 \\
    OCE-Net \cite{wei2023orientation} & 6.30 & 83.02 & 72.22 & 80.18 & 98.12 & 83.41 & 73.67 & 80.62 & 98.72 & 81.96 & 69.87 & 81.38 & 98.24 \\
    MAGF-Net \cite{li2023magf}      & 34.60 & 83.07 & 70.32 & 82.62 & 98.62 & 83.64 & 74.65 & 84.84 & 96.49 & 81.96 & 72.43 & 81.38 & 98.95 \\
    DCNet \cite{Shang2024}          & 1.05  & 82.94 & 71.95 & 82.08 & 98.02 & 82.50 & 71.35 & 82.30 & 96.77 & 83.32 & 72.76 & 82.05 & 98.17 \\
    FS-UNet \cite{jiang2024retinal} & 0.87  & 82.46 & 70.19 & 80.71 & \textbf{98.60} & 84.05 & 72.67 & 83.80 & 98.05 & 81.33 & 68.56 & 83.42 & 98.53 \\
    G-Net Light \cite{iqbal2022g}   & 0.39  & 82.02 & 69.09 & 81.92 & 98.29 & 82.78 & 69.64 & 81.70 & 98.53 & 80.48 & 67.76 & 82.10 & 98.38 \\
    LMBiS-Net \cite{Abbasi2024}     & 0.17  & 83.43 & 72.65 & 82.60 & 98.07 & 84.44 & 74.64 & 84.73 & 98.44 & 83.54 & 73.76 & 83.05 & 98.16 \\
    \textbf{LFRA-Net (ours)}         & \textbf{0.17} & \textbf{84.28} & \textbf{72.86} & 82.43 & 98.08 & \textbf{88.44} & \textbf{79.31} & \textbf{88.75} & 98.56 & \textbf{85.50} & \textbf{74.70} & 84.36 & 98.20 \\
    \bottomrule
\end{tabular}
}
\end{table*}
\textbf{Implementation Details}
We use TensorFlow and Keras to create our model using an NVIDIA RTX A4000 with 16 GB of GDDR6 VRAM. To improve generalization and mitigate overfitting due to limited dataset size, we applied data augmentation during training. It is performed by rotating the images by 20 degrees to include modifications in acquisition angles and modifying their contrast to include modifications in lighting and image quality. We applied primary augmentation techniques to all three datasets, namely CLoDSA\footnote{\url{https://github.com/joheras/CLoDSA}{https://github.com/joheras/CLoDSA}} and IMGAUG\footnote{\url{https://github.com/aleju/imgaug}{https://github.com/aleju/imgaug}}.

% The model is first trained on a combined set of images from the three datasets with data augmentation to cope with minor dataset issues in medical images and then fine-tuned on each dataset using the weights from the combined training, resulting in improved performance.
\subsection{Results}\label{subsec:comparisionsBV}
\begin{figure*}[!h]
  \centering
    \includegraphics[width=\textwidth]{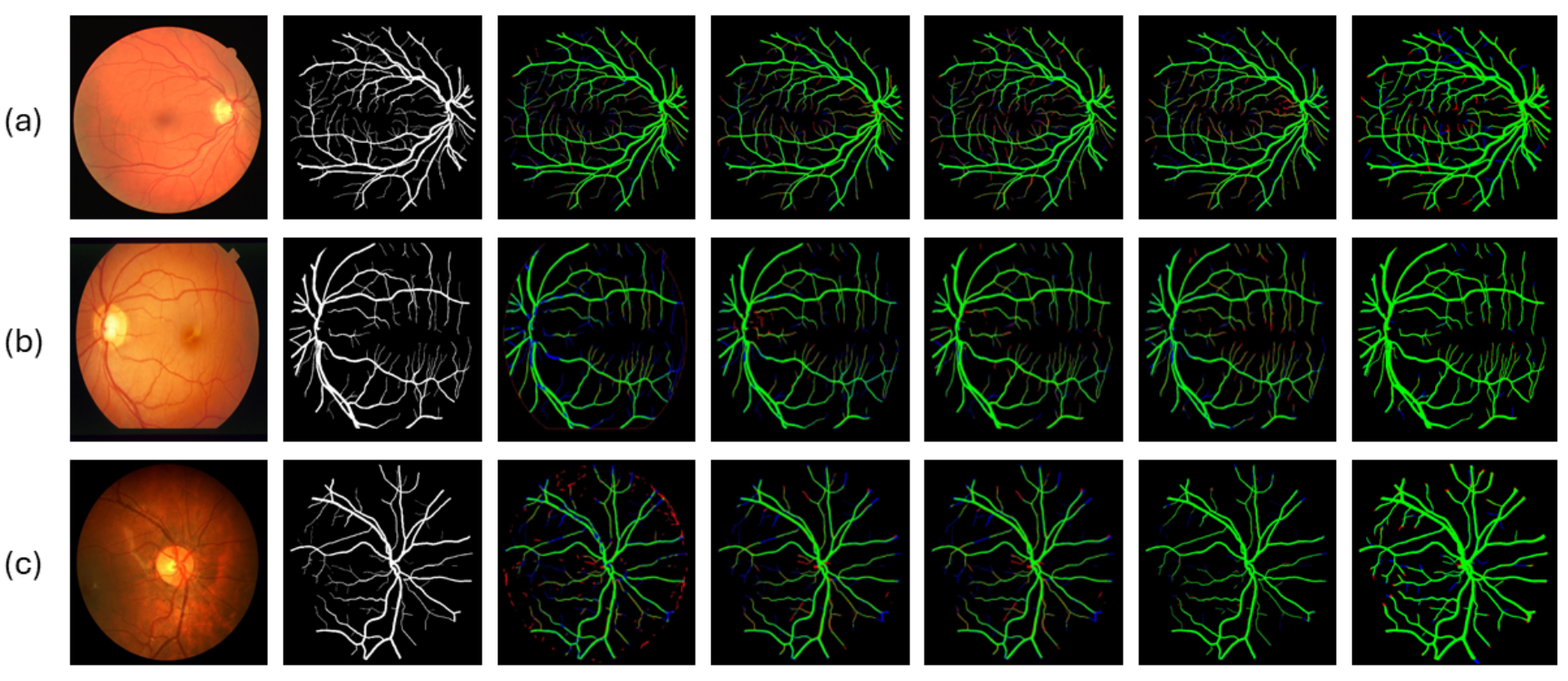}
    \caption{Segmentation outcomes of retinal vessel segmentation for three datasets: (a) DRIVE, (b) STARE and (c) CHASE\_DB. The images displayed in the following order are input images, the ground truth and the outputs of the G-Net Light, MultiResNet, SegNet, U-Net++, and LFRA-Net models.}\label{fig:combined}
\end{figure*}

A computational complexity and comprehensive evaluation of LFRA-Net performance compared to the existing state-of-the-art methods on the DRIVE, STARE and CHASE\_DB datasets are presented in Table \ref{complex_results} and Table \ref{tab:Vessels}, respectively. These results demonstrate the effectiveness of LFRA-Net in segmenting retinal vessels, ensuring fewer false negatives and accurate segmentation. One of the primary characteristics of LFRA-Net is its computational efficiency. Compared to other models, it is incredibly lightweight, with only 0.17 million parameters. It exhibits a well-balanced trade-off between sensitivity and specificity, sustaining robust performance in both metrics. Despite having the same parameter size, LFRA-Net outperforms LMBiS-Net \cite{Abbasi2024} across all datasets. Moreover, visual representation of segmentation quality offers valuable insights into the model’s ability to detect fine vessel structures. Figure \ref{fig:combined} shows the visual representation of the segmentation results for various existing models on all three datasets. Each model's performance is illustrated by color-coded overlays: blue pixels indicate false positive detections, and green pixels indicate accurate positive detections. In the images, LFRA-Net generates the most accurate segmentations among the models, with the fewest false positives and false negatives.

\subsection{Ablation Study on DRIVE dataset}
\begin{table*}[!h]
  \centering
  \caption{Ablation study of the proposed model with modifications to evaluate the impact of various components on the baseline model.}
   \resizebox{\textwidth}{!}{%
    \begin{tabular}{lcccccc}
    \toprule
    \multirow{2}[4]{*}{\textbf{Method}} & \multicolumn{6}{c}{\textbf{Performance Measures (\%)}} \\
    \cmidrule{2-7} 
        & \textbf{Param (M)}  & \textbf{Dice} & \textbf{J} & \textbf{Acc} & \textbf{Sen} & \textbf{Sp} \\
    \midrule
       Lightweight UNet no-skip connection (LU-NS) & 0.07 & 80.30 & 64.11 & 93.65 & 69.02 & 96.93 \\
        Multiscale LU no skip connection (MLU-NS) (1\,$\times$\,1, 3\,$\times$\,3, Dilated 3\,$\times$\,3) & 0.10 & 80.92 & 64.01 & 94.62 & 70.10 & 97.10 \\
        Multiscale LU skip connection (MLU) (1\,$\times$\,1, 3\,$\times$\,3, Dilated 3\,$\times$\,3) & 0.10 & 81.32 & 66.99 & 95.66 & 71.47 & 97.43 \\
        MLU + ($\mathcal{F}$) in skip connections (($\mathcal{F}$)-Skip) & 0.15 & 82.25 & 69.23 & 95.96 & 81.48 & 97.42 \\
        MLU + ($\mathcal{R}$) in skip connections (($\mathcal{R}$)-Skip) & 0.16 & 82.35 & 70.49 & 95.83 & 73.99 & 97.86 \\
        MLU + ($\mathcal{R}$)-Skip + ($\mathcal{F}$) in Bottleneck ($\mathcal{F}$)-Bottleneck) & 0.18 & 83.77 & 71.31 & 95.99 & 77.40 & 98.05 \\
        MLU + ($\mathcal{F}$) in Bottleneck (($\mathcal{F}$)-Bottleneck) & 0.15 & 83.61 & 71.11 & 96.07 & 80.18 & 97.77 \\
        MLU + ($\mathcal{R}$) in 1\,3\,-Skip + ($\mathcal{F}$)-Bottleneck & 0.17 & 83.63 & 71.89& 95.98 & 80.71 &	98.02\\
        MLU + ($\mathcal{R}$) in 2\,3\,-Skip + ($\mathcal{F}$)-Bottleneck & 0.19 & 83.64 & 71.90 & 95.89 & 82.42 & 97.86 \\
       \textbf{ MLU + ($\mathcal{R}$) in 1\,2\,-Skip + ($\mathcal{F}$)-Bottleneck }& 0.17 & \textbf{84.28} & \textbf{72.86} & \textbf{96.09} & \textbf{82.43} & \textbf{98.09} \\
    \bottomrule
    \end{tabular}%
    }
  \label{tab:Ablation}%
\end{table*}%

The ablation study, presented in Table \ref{tab:Ablation}, illustrates the effectiveness of our segmentation models with some alterations and configurations. The baseline is the lightweight UNet without skip connections (LU-NS), which has a dice of 80.30\%. It is slightly improved by adding multiscale features without skip connections (MLU-NS). By incorporating skip connections into the multiscale LU model (MLU), specificity is further enhanced. Nevertheless, the dice and sensitivity rise when the region-aware attention mechanism ($\mathcal{R}$) is added. Adding focal modulation ($\mathcal{F}$) to bottlenecks results in the highest jaccard and dice.
The optimal overall segmentation performance is improved by integrating $\mathcal{F}$ and $\mathcal{R}$ s in the first and second skip connections and bottleneck, respectively. LFRA-Net is the first lightweight encoder-decoder to strategically combine region-aware attention in early skip connections with focal modulation attention at the bottleneck. This careful placement retains precise spatial details while capturing global context, resulting in state-of-the-art accuracy with only 0.17 million parameters.

\section{Conclusion}
In this research, we presented LFRA-Net, a novel lightweight multiscale encoder-decoder network for retinal blood vessel segmentation designed to address the issues of existing methods in capturing small vessel structures and depending on computationally expensive architectures. Our method uses advanced attention mechanisms, such as region-aware attention in the selective skip connections and focal modulation attention in the encoder-decoder bottleneck, to capture local and global dependency, ensuring robust segmentation performance. Comprehensive tests on the DRIVE, STARE, and CHASE\_DB datasets show that LFRA-Net performs satisfactorily on several measures, such as accuracy, sensitivity, Jaccard index, and Dice score, with only 0.17 million parameters. The combination of region-aware attention in early skip connection and focal modulation at the bottleneck is key to the balance of accuracy and computational efficiency of LFRA-Net, enabling a model to outperform heavy models. This shows that LFRA-Net is suitable for retinal image segmentation applications with real-time processing and limited resources.

\newpage

\end{document}